\documentclass{article}

\usepackage{PRIMEarxiv}

\usepackage[hyphens]{url}
\usepackage[utf8]{inputenc} 
\usepackage[T1]{fontenc}    
\usepackage{hyperref}       
\usepackage{booktabs}       
\usepackage{amsfonts}       
\usepackage{nicefrac}       
\usepackage{microtype}      
\usepackage{lipsum}
\usepackage{fancyhdr}       
\usepackage{graphicx}       
\graphicspath{{Figures/}}     
\usepackage{longtable}

\usepackage{diagbox}

\title{Generalist embedding models are better at short-context clinical semantic search than specialized embedding models}

\author{
  \normalfont
  \begin{tabular}{>{\raggedright\arraybackslash}p{0.9\textwidth}}
    \textbf{Jean-Baptiste Excoffier\textsuperscript{1}}\thanks{\textbf{Corresponding Author}: \texttt{jeanbaptiste.excoffier@kaduceo.com} }  ,
    \textbf{Tom Roehr\textsuperscript{2}},
    \textbf{Alexei Figueroa\textsuperscript{2}},
    \textbf{Jens-Michalis Papaioannou\textsuperscript{2}},
    \textbf{Keno Bressem\textsuperscript{3}},
    \textbf{Matthieu Ortala\textsuperscript{1}}\\
    \\
     \textsuperscript{1}Kaduceo, Toulouse, France \\
     \textsuperscript{2}Berliner Hochschule fur Technik, Berlin, Germany \\
    \textsuperscript{3}Department of Radiology and Nuclear Medicine, German Heart Center Munich, Munich, Germany \\
    \\
  \end{tabular}
}

\clearpage

\begin{document}

\maketitle

\begin{abstract}
The increasing use of tools and solutions based on Large Language Models (LLMs) for various tasks in the medical domain has become a prominent trend. Their use in this highly critical and sensitive domain has thus raised important questions about their robustness, especially in response to variations in input, and the reliability of the generated outputs.
This study addresses these questions by constructing a textual dataset based on the ICD-10-CM code descriptions, widely used in US hospitals and containing many clinical terms, and their easily reproducible rephrasing. We then benchmarked existing embedding models, either generalist or specialized in the clinical domain, in a semantic search task where the goal was to correctly match the rephrased text to the original description.
Our results showed that generalist models performed better than clinical models, suggesting that existing clinical specialized models are more sensitive to small changes in input that confuse them. The highlighted problem of specialized models may be due to the fact that they have not been trained on sufficient data, and in particular on datasets that are not diverse enough to have a reliable global language understanding, which is still necessary for accurate handling of medical documents.
\end{abstract}

\bigskip


\clearpage

\section{Motivations}%
\label{Motivations}

In recent years, the use of Large Language Models (LLMs) in the medical domain has increased significantly \cite{Moor2023, Qiu2023, Omiye2023, Shoja2023}. They are used for multiple and diverse tasks such as chatbots \cite{Kim2023}, document summarization \cite{Singhal2023}, anonymization, and identification of precise clinical information such as automatic recognition of medical codes (including ICD-10 and CPT billing codes) \cite{Moor2023, Thirunavukarasu2023}.\\
These models are often components of broader systems and solutions that contribute to the consolidation and improvement of the final results by incorporating multiple prompts from different LLMs, as well as the integration of technologies such as Retrieval-Augmented Generation (RAG \cite{Lewis2020}) and semantic search \cite{Tamine2021}.

The practical deployment of LLMs in the highly sensitive medical field raises critical questions about their robustness for daily use in a real medical environment. Ensuring the reliability of results are crucial concerns, in particular, developing a comprehensive understanding of the model's behavior, preventing hallucination and incoherence in the generated content with a focus on the medical soundness of the outputs \cite{Ji2023, Xie2023}, as well as limiting the sensitivity of prompt formatting and input, which has been observed to be high \cite{Sclar2023}. Addressing these issues is critical for establishing the trustworthiness of LLM applications in medical contexts \cite{Omiye2023, Harrer2023, Wang2023, He2023}.\\
In particular, the effectiveness of embedding models in medical semantic search tasks is an area that requires further exploration \cite{Huang2022, Tamine2021}. Identifying the types of models that yield optimal performance and understanding the underlying reasons for their effectiveness are essential steps toward refining and optimizing search capabilities within medical datasets that are useful for multiple tasks such as document comparison and retrieval or medical information retrieval.

Our study addresses such current and pressing concerns about the robustness and effectiveness of LLMs in the medical domain through a comprehensive benchmark on semantic search tasks in the clinical domain. This benchmark aims to evaluate the performance of different embedding models to efficiently retrieve short clinical diagnostic information, even when rephrased and reformulated, as is always the case in human-written medical documents such as discharge notes. A rigorous benchmarking experiment will not only shed light on the capabilities of different models, but will also provide valuable insights into their comparative strengths and weaknesses.\\
Reproducibility of benchmarking experiments is essential in this study. The proposed benchmarking framework is therefore designed to be easily replicable, allowing research teams with limited resources to contribute to and build upon the results. This commitment to reproducibility serves as a foundation for future collaborative efforts and the collective advancement of the field.

\section{Methodology}%
\label{Methodology}

\subsection{Generated dataset}%
We used the well-known ICD-10-CM codes, which consist of codes of 3 to 7 characters (digits and text characters) that correspond to medical diagnoses \cite{Hirsch2016}. The ICD-10-CM classification system is the most widely used in the U.S. for hospital billing since the mid-2010s \cite{Sanders2012}.

Each code has a description (e.g., code O30 has "Multiple gestation" as a description) that can be embedded, i.e., transformed into a numeric vector, using an embedding model. The same embedding model can embed a new phrase (e.g., "Multiple pregancies") and associate it with the nearest embedded ICD-10-CM description using a given distance function. Finally, the code corresponding to the closest description can be associated with the new sentence. The quality of this semantic search procedure is highly dependent on the intrinsic quality of the embedding model to accurately embed medical text in order to retrieve the correct ICD-10-CM code.

For this purpose, we built a dataset of 100 ICD-10-CM codes along with their description and ten reformulations for each. The full dataset with the selected ICD-10-CM codes and their reformulations was made public and is available at \footnote{\url{https://github.com/kaduceo/ICD10CM_Embedding_Benchmark}}. Here is the detailed procedure for building the dataset:
\begin{enumerate}
  \item Random selection of 100 ICD-10-CM codes. 50 of these were categories (i.e., codes with only three characters) and 50 were six-character codes.
  \item Using OpenAI ChatGPT 3.5 turbo model via Python API \footnote{OpenAI API for Python language : \url{https://github.com/openai/openai-python} and ChatGPT 3.5 Turbo release and update notes : \url{https://openai.com/blog/gpt-3-5-turbo-fine-tuning-and-api-updates}.}, the description of each code was reformulated 10 times. We made sure that all reformulations were different from the original description and that there were no duplicates.
  \item The final size of the dataset was therefore 1000 items, with 100 different codes and 1000 different reformulations.
\end{enumerate}

Table \ref{single_complete_example_of_reformulation_process} shows an example of a code (S48) along with the 10 reformulations of its original description produced by the generative model.

\begin{table}[!ht]%
	\caption{Complete example of a code and its 10 reformulations produced by the generative LLM.}%
	\centering%
	\begin{tabular}{|p{30mm}|p{85mm}|}
		\hline
        ICD-10-CM code & S48\\[0.15cm]\hline
	Original description & Traumatic amputation of shoulder and upper arm\\[0.15cm]\hline
        Reformulation 1 & Traumatic amputation of the shoulder and upper arm\\[0.15cm]\hline
        Reformulation 2 & Shoulder and upper arm traumatic amputation\\[0.15cm]\hline
        Reformulation 3 & Amputation of shoulder and upper arm due to trauma\\[0.15cm]\hline
        Reformulation 4 & Upper arm and shoulder traumatic amputation\\[0.15cm]\hline
        Reformulation 5 & Traumatically amputated shoulder and upper arm\\[0.15cm]\hline
        Reformulation 6 & Shoulder and upper arm amputated due to trauma\\[0.15cm]\hline
        Reformulation 7 & Traumatic removal of shoulder and upper arm\\[0.15cm]\hline
        Reformulation 8 & Amputation of the upper arm and shoulder caused by trauma\\[0.15cm]\hline
        Reformulation 9 & Traumatic dismemberment of the shoulder and upper arm\\[0.15cm]\hline
        Reformulation 10 & Shoulder and upper arm dismemberment as a result of trauma\\[0.15cm]\hline
	\end{tabular}
	\label{single_complete_example_of_reformulation_process}%
\end{table}

\subsection{Semantic search}%
Given an embedding model, we fully embedded the ICD-10-CM code descriptions and used them as retrieval corpus. We then embedded the reformulation dataset using the same model to use them as queries. For each query, we retrieved the top one code description using Euclidean distance. 
We evaluated the performance by comparing the retrieved code description and associated code with the code associated with the reformulated code description.

We considered a "good" embedding for this clinical task as being able to retrieve the correct (i.e., original) code, even if it was only passed as a reformulation.\\
Inaccurate reformulations, in the sense that they are too far from the original description to prevent any embedding model from retrieving that original description, were not considered a problem since they are rare and, more importantly, all benchmarked embedding models would be wrong, so it would not affect the final comparison between metric results.

\subsection{Embedding models}
We used both generalist and clinical embedding models based on two conditions. The first is that they must be cheap to derive in the sense that they do not require a GPU to run but only a CPU (8GB), with the intention that the experiment is easily and quickly reproducible by anyone without extensive computing resources. The second condition is that we used free to use (thus excluding paid APIs such as OpenAI) and widely used models (based on the number of downloads on the HuggingFace website \footnote{HuggingFace website : : \url{https://huggingface.co/}.} \cite{Jain2022}), still with the same goal of allowing anyone to reproduce our experiments for free and with what are considered to be well-known and robust models.

Table \ref{embedding_models_details} details all 19 embedding models used in this study. The S-PubMedBERT embedding model is based on a clinical model (PubMedBERT) that has been further trained on a larger generalist dataset (MS-MARCO \cite{Deka2022}), so S-PubMedBERT was considered as a generalist embedding model in this study.

\clearpage

\begin{table}
\caption{Detailed information of embedding models used in this study.}%
\centering%
\begin{tabular}{|p{24mm}|p{18mm}|p{16mm}|p{60mm}|p{16mm}|}
\hline%
  \textbf{Name} & \textbf{Is health specialized?} & \textbf{Paper (if available)} & \textbf{HuggingFace page} & \textbf{Embedding vector size}\\[0.15cm]\hline
  e5-small-v2 & No & \cite{Wang2022} & \url{https://huggingface.co/intfloat/e5-small-v2}  & 384\\[0.15cm]\hline
	 e5-large-v2 & No & \cite{Wang2022} & \url{https://huggingface.co/intfloat/e5-large-v2}  & 1024 \\[0.15cm]\hline
	 all-MiniLM-L6-v2 & No & \cite{Wang2020} & \url{https://huggingface.co/sentence-transformers/all-MiniLM-L6-v2} & 384\\[0.15cm]\hline
	 all-MiniLM-L12-v2 & No & \cite{Wang2020} & \url{https://huggingface.co/sentence-transformers/all-MiniLM-L12-v2}  & 384\\[0.15cm]\hline
	 bert-base-uncased & No & \cite{Devlin2019} & \url{https://huggingface.co/bert-base-uncased}& 768\\[0.15cm]\hline
	 bert-large-uncased & No & \cite{Devlin2019} & \url{https://huggingface.co/bert-large-uncased}  & 1024\\[0.15cm]\hline
	 longformer-base-4096 & No & \cite{Beltagy2020} & \url{https://huggingface.co/allenai/longformer-base-4096}  & 768\\[0.15cm]\hline
	 jina-embeddings-v2-base-en & No & \cite{Gunther2023} & \url{https://huggingface.co/jinaai/jina-embeddings-v2-base-en} & 768\\[0.15cm]\hline
	 bigbird-roberta-base & No & \cite{Zaheer2020} & \url{https://huggingface.co/google/bigbird-roberta-base} &768\\[0.15cm]\hline
    S-PubMedBERT & No & \cite{Deka2022} & \url{https://huggingface.co/pritamdeka/S-PubMedBert-MS-MARCO}  & 768\\[0.15cm]\hline
	 MedBERT & Yes & \cite{Rasmy2021} & \url{https://huggingface.co/Charangan/MedBERT} & 768\\[0.15cm]\hline
	 ClinicalBERT & Yes & \cite{Huang2019} & \url{https://huggingface.co/medicalai/ClinicalBERT} & 768\\[0.15cm]\hline
	 BioBERT & Yes & \cite{Lee2020} & \url{https://huggingface.co/dmis-lab/biobert-v1.1} &  768\\[0.15cm]\hline
	 CORe-clinical-outcome-BioBERT  & Yes & \cite{VanAken2021} & \url{https://huggingface.co/bvanaken/CORe-clinical-outcome-biobert-v1} & 768\\[0.15cm]\hline
    PubMedBERT & Yes & \cite{Gu2021} & \url{https://huggingface.co/microsoft/BiomedNLP-BiomedBERT-base-uncased-abstract-fulltext}  & 768\\[0.15cm]\hline
	 Clinical-Longformer & Yes & \cite{Li2022} & \url{https://huggingface.co/yikuan8/Clinical-Longformer} & 768\\[0.15cm]\hline
	 Clinical-BigBird & Yes & \cite{Li2022} & \url{https://huggingface.co/yikuan8/Clinical-BigBird} &768\\[0.15cm]\hline
	 Medical-T5-Large & Yes & No published paper yet & \url{https://huggingface.co/Falconsai/medical_summarization} & 512\\[0.15cm]\hline
	 GatortronS & Yes & \cite{Peng2023} & \url{https://huggingface.co/UFNLP/gatortronS} &  1024\\[0.15cm]\hline
  \end{tabular}
\label{embedding_models_details}%
\end{table}

\subsection{Metrics}
For each embedding and reformulation, we compared the retrieved code to the original using the following metrics. Table \ref{metrics_examples} shows the metrics results for four manual examples.
\begin{itemize}
\setlength\itemsep{-0.2em}
	\item \textit{Exact matching}: if the retrieved code is exactly the same as the original.
	\item \textit{Category matching}: if the category of the retrieved code is the same as the category of the original code. The category is defined by the first three characters of the code.
	\item Character error rate (CER) between the retrieved code and the original code. 
\end{itemize}

\begin{table}[!ht]%
	\caption{Metrics results for four examples.}%
	\centering%
	\begin{tabular}{|p{17mm}|p{17mm}|p{17mm}|p{17mm}|p{17mm}|}
		\hline
        \textbf{Original code} & \textbf{Retrieved code} & \textit{Exact matching} & \textit{Category matching} & \textit{CER}\\[0.15cm]\hline
        E28319 & E28319 & Correct & Correct & 0\\[0.15cm]\hline
        O30 & O301 & Incorrect & Correct & 0.33\\[0.15cm]\hline
        E28319 & E2831 & Incorrect & Correct & 0.16\\[0.15cm]\hline
        O30 & O312 & Incorrect & Incorrect & 0.66\\[0.15cm]\hline
	\end{tabular}
	\label{metrics_examples}%
\end{table}

The final metrics are presented using the average of all reformulations. As for the CER metrics, we distinguished between the total CER (\textit{Total CER}), computed over all reformulations, and the incorrect CER (\textit{Incorrect CER}), computed only on the incorrectly retrieved codes (different for each embedding model). The latter allows a better view of the level of accuracy even if the model is incorrect (in the sense of exact code matching).

\section{Results}%
\label{Results}

\subsection{Performance Metrics}
The embedding model with the highest performance was jina-embeddings-v2-base-en, except for the Incorrect CER metrics where all-MiniLM-L12-v2 had the best (i.e. lowest) value. Both jina-embeddings-v2-base-en and all-MiniLM-L12-v2 are generalist models, while the best clinical model was ClinicalBERT. Detailed results for all embedding models are shown in table \ref{embedding_models_results}.\\
There was a large gap between the generalist and clinical models for all metrics. The top generalist model (jina-embeddings-v2-base-en) had an Exact Matching rate of 84.0\% compared to only 64.4\% for the top clinical model (ClinicalBERT). Furthermore, the generalist models were also closer to the ground truth than the clinical models, even when they did not retrieve the exact code (Incorrect CER metrics).

\begin{longtable}{ |p{50mm}|p{20mm}|p{20mm}|p{18mm}|p{18mm}| }
\caption{Detailed performance results for each embedding model.\\ All results are displayed in \%. Performances are the highest when \textbf{(1)} metrics are the highest and \textbf{(2)} metrics are the lowest.}\\%
\multicolumn{5}{l}{\label{embedding_models_results}}\\%
\hline%
  \textbf{Embedding model} & \textbf{Exact Matching (1)} & \textbf{Category Matching (1)} & \textbf{Total CER (2)} & \textbf{Incorrect CER (2)}\\[0.15cm]\hline
  e5-small-v2 & 82.6 & 92.1 & 10.93 & 57.6\\[0.15cm]\hline
	 e5-large-v2 & 81.3 & 92.0 & 10.6 & 56.8\\[0.15cm]\hline
	 all-MiniLM-L6-v2 & 76.2 & 91.0 & 12.9 & 54.1\\[0.15cm]\hline
	 all-MiniLM-L12-v2 & 74.6 & 92.5 & 12.6 & \textbf{49.7}\\[0.15cm]\hline
	 bert-base-uncased & 52.1 & 76.4 & 30.0 & 62.7\\[0.15cm]\hline
	 bert-large-uncased & 31.5 & 65.1 & 45.1 & 65.8\\[0.15cm]\hline
	 longformer-base-4096 & 48.6 & 73.2 & 34.8 & 67.6\\[0.15cm]\hline
	 jina-embeddings-v2-base-en & \textbf{84.0} & \textbf{92.8} & \textbf{9.2} & 57.6\\[0.15cm]\hline
	 bigbird-roberta-base & 36.0 & 64.2 & 44.7 & 69.8\\[0.15cm]\hline
     S-PubMedBERT & 78.1 & 91.8 & 12.4 & 56.7\\[0.15cm]\hline
	 MedBERT & 44.8 & 70.4 & 39.1 & 70.9\\[0.15cm]\hline
	 ClinicalBERT & 64.4 & 82.6 & 25.45 & 71.5\\[0.15cm]\hline
	 BioBERT & 50.5 & 78.8 & 30.9 & 62.5\\[0.15cm]\hline
	 CORe-clinical-outcome-BioBERT & 53.1 & 80.3 & 28.7 & 61.2\\[0.15cm]\hline
    PubMedBERT & 48.9 & 76.0 & 34.0 & 66.5\\[0.15cm]\hline
	 Clinical-Longformer & 46.9 & 77.1 & 34.0 & 64.1\\[0.15cm]\hline
	 Clinical-BigBird & 38.2 & 69.3 & 43.0 & 69.6\\[0.15cm]\hline
	 Medical-T5-Large & 52.6 & 73.9 & 35.8 & 75.5\\[0.15cm]\hline
	 GatortronS & 41.1 & 76.9 & 31.8 & 53.9\\[0.15cm]\hline
\end{longtable}

\subsection{Visual performance}
\begin{figure}[!ht]
    \centering
    \begin{minipage}{0.48\textwidth}
        \centering
        \includegraphics[scale=0.4]{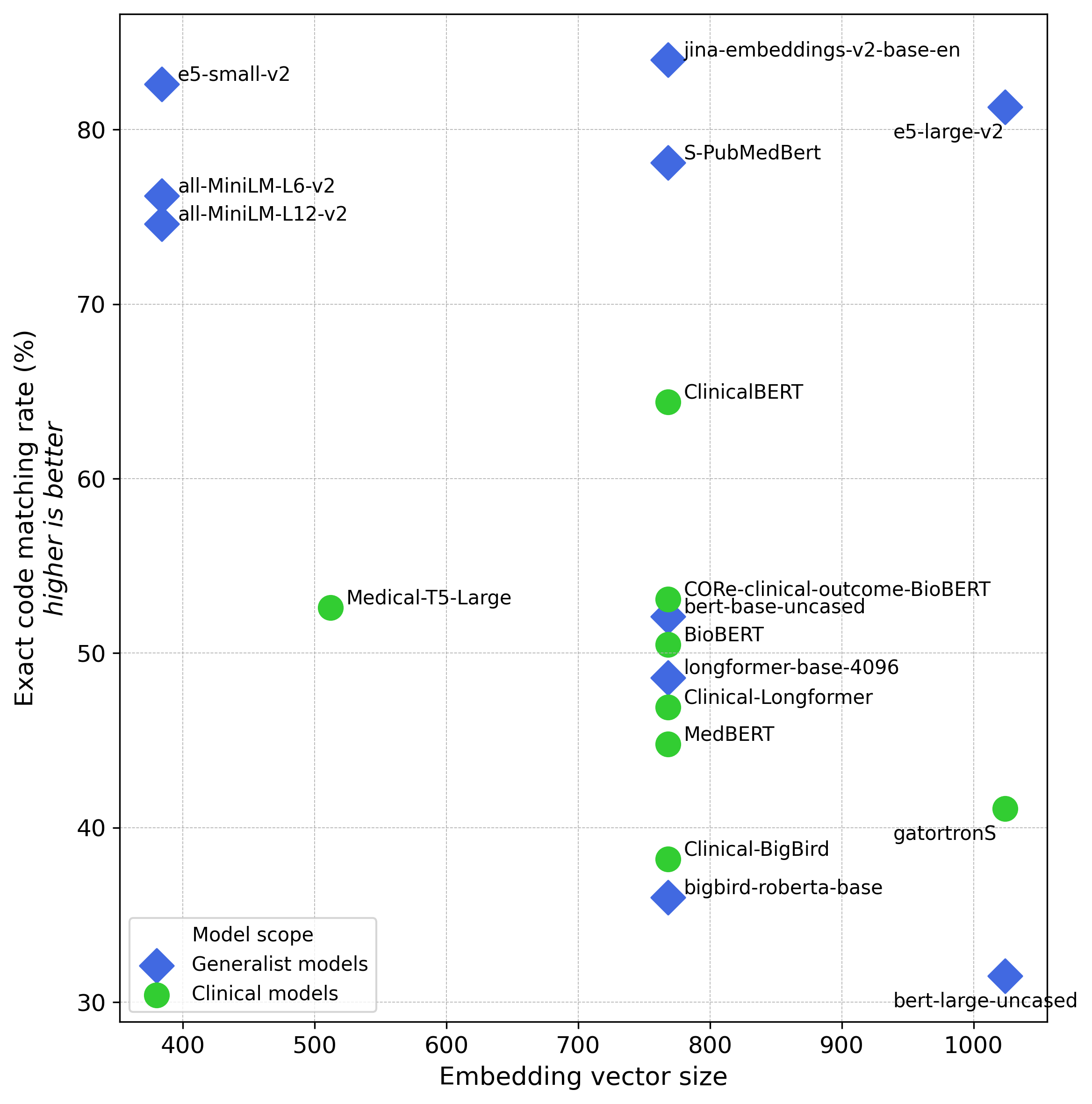}
        \caption{Exact code retrieval performances depending on model scope and embedding size.\\}
        \label{embedding_size_vs_exact_code_matching}
    \end{minipage}\hfill
    \begin{minipage}{0.48\textwidth}
        \centering
        \includegraphics[scale=0.4]{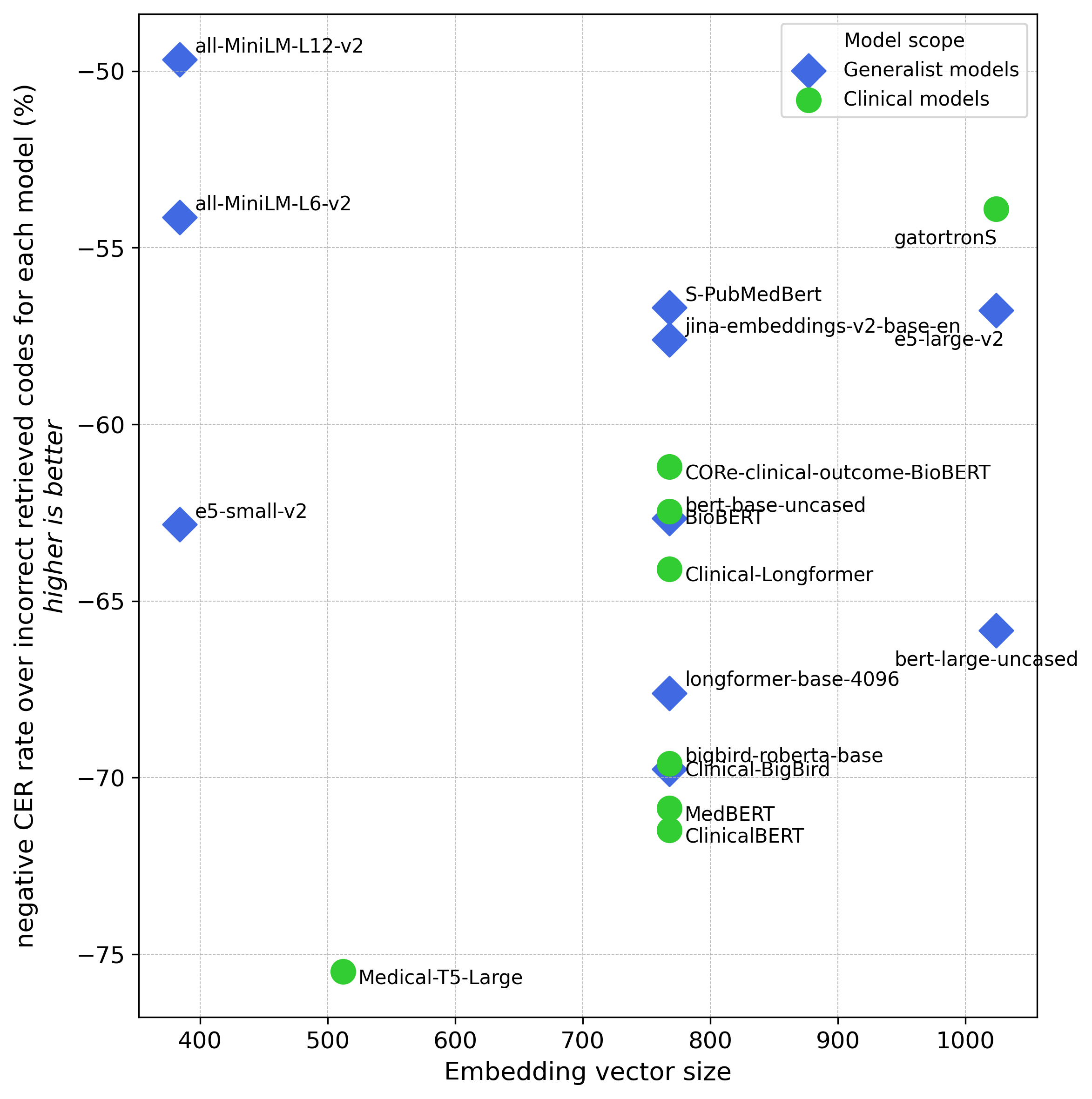}
        \caption{Negative CER rate of incorrect retrieved code performances depending on model scope and embedding size.}
        \label{embedding_size_vs_negative_CER_unmatched}
    \end{minipage}
\end{figure}

Figures \ref{embedding_size_vs_exact_code_matching} and \ref{embedding_size_vs_negative_CER_unmatched} present 3D plots of the embedding models' performance with respect to their scope (generalist or specialized in the clinical domain) and embedding vector size. A smaller embedding vector size (less than or equal to 600) was associated with higher performance, either for the exact matching rate or for the incorrect CER.\\
The jina-embeddings-v2-base-en and the two versions of the e5-v2 model (small and large) had the highest exact matching rate, as shown in Figure \ref{embedding_size_vs_exact_code_matching}. However, the top three models with respect to the Incorrect CER metrics were the two versions of the all-MiniLM-v2 model (L6 and L12) and the GatortronS, as shown in Figure \ref{embedding_size_vs_negative_CER_unmatched}. 

\subsection{Selected examples}
We called the \textit{TOP 3 GENERAL} the 3 embedding models without specialization with the highest exact rates : jina-embeddings-v2-base-en, e5-small-v2 and e5-large-v2.\\
We called the \textit{TOP 3 MEDICAL} the 3 embedding models with specialization in health with the highest exact rates : ClinicalBERT, CORe-clinical-outcome-BioBERT and Medical-T5-Large.\\
For each combination of \textit{TOP 3 GENERAL} and \textit{TOP 3 MEDICAL} producing correct or incorrect codes, 2 random examples are presented in Table \ref{selected_examples}. n indicates the total number of observations (i.e. reformulations) belonging to the combination,  for a total of 1000 observations as described in the Methodology section.

\begin{table}[!ht]
\caption{Examples of model responses (i.e. code retrieval), with same category indicated in bold.}%
\centering%
\begin{tabular}{|p{12mm}|p{18mm}|p{18mm}|p{12mm}|p{12mm}|p{12mm}|p{12mm}|p{12mm}|p{12mm}|}
\hline
\multicolumn{3}{|c}{Code infos} & \multicolumn{3}{|c}{\textit{TOP 3 GENERAL} code retrieval} & \multicolumn{3}{|c|}{\textit{TOP 3 MEDICAL} code retrieval} \\\hline
Code & Description & Reformula-tion & jina-embedd ings-v2-base-en & e5-small-v2 & e5-large-v2 & Clinical BERT & CORe-clinical-outcome-BioBERT & Medical-T5-Large \\\hline
\multicolumn{3}{|c}{$n = 354$} & \multicolumn{3}{|c}{Correct} & \multicolumn{3}{|c|}{Incorrect} \\\hline
S48019 & Complete traumatic amputation at unspecified shoulder joint & Unspecified complete shoulder joint traumatic amputation & \textbf{S48019} & \textbf{S48019} & \textbf{S48019} & \textbf{S48019} & \textbf{S48019} & \textbf{S48019}\\\hline
S97 & Crushing injury of ankle and foot & Ankle and foot crushing injury & \textbf{S97} & \textbf{S97} & \textbf{S97} & \textbf{S97} & \textbf{S97} & \textbf{S97}\\\hline
\multicolumn{3}{|c}{$n = 91$} & \multicolumn{3}{|c}{Incorrect} & \multicolumn{3}{|c|}{Incorrect} \\\hline
O00 & Ectopic pregnancy & Pregnancy in the wrong place & Z331 & O9821 & Z640 & Z98870 & T191 & O26813\\\hline
E28319 & Asymptomatic premature menopause & Early menopause without any symptoms & \textbf{E2831} & \textbf{E2831} & Z780 & N959 & \textbf{E2831} & \textbf{E28310} \\\hline
\multicolumn{3}{|c}{$n = 87$} & \multicolumn{3}{|c}{Correct} & \multicolumn{3}{|c|}{Incorrect} \\\hline
I50 & Heart failure & Heart deterioration & \textbf{I50} & \textbf{I50} & \textbf{I50} & T551 & \textbf{I5081} & O76\\\hline
O00112 & Left tubal pregnancy with intrauterine pregnancy & IUP with concurrent ectopic pregnancy in the left tube & \textbf{O00112} & \textbf{O00112} & \textbf{O00112} & O083 & O083 & M84752K \\\hline
\multicolumn{3}{|c}{$n = 2$} & \multicolumn{3}{|c}{Incorrect} & \multicolumn{3}{|c|}{Correct} \\\hline
Z30 & Encounter for contraceptive management & Conference for contraceptive oversight & \textbf{Z304} & \textbf{Z304} & \textbf{Z304} & \textbf{Z30} & \textbf{Z30} & \textbf{Z30}\\\hline
I50 & Heart failure & Heart collapse & J981 & J981 & J981 & \textbf{I50} & \textbf{I50} & \textbf{I50}\\\hline
\end{tabular}
\label{selected_examples}%
\end{table}

\clearpage

\section{Discussion}%
\label{Discussion}

As shown in table \ref{embedding_models_results}, the best performing models for short-context clinical semantic search are generalist models, with the top 3 models being jina-embeddings-v2-base-en, e5-small-v2 and e5-large-v2.\\
The 3 best clinical embedding models for this task are ClinicalBERT and CORe-clinical-outcome-BioBERT, both based on the BERT architecture, and Medical-T5-Large. However, the best generalist models outperformed the single best specialist model (ClinicalBERT) by a wide margin of 15 to 20\% for all metrics. And it is important to note that ClinicalBERT itself outperformed other clinically oriented models as well by a comparable margin (10 to 20\%).\\

These results were a priori unexpected. We would have expected clinical models to perform better than generalist models, since our experiment consists of rephrasing and semantic retrieval of medical diagnosis descriptions (ICD-10-CM codes), which contain a large number of precise medical and clinical terms \cite{Kaur2021}.\\
The context of this experiment, consisting of small sentences, favored the sentence-transformer models, with all 3 best performing embedding models being both generalist and sentence-transformer. This gap was highlighted by the direct comparison of PubMedBERT and S-PubMedBERT results, with S-PubMedBERT being 29\% more precise than PubMedBERT on the exact matching rate metrics. In fact, PubMedBERT is a fully clinical embedding model \cite{Gu2021}, while S-PubMedBERT used PubMedBERT as a basis for further sentence transformer training on a larger general dataset \cite{Deka2022}. However, S-PubMedBERT was still outperformed by several fully generalized sentence-transformer embedding models (jina-embeddings-v2-base-en and the two e2-v2 models).\\

Thus, our experiment highlights two main points. The first is the need for an appropriate training phase that matches the final needs. In our case, this was indicated by the fact that the best embedding models were all sentence transformers, as well as by the superiority of S-PubMedBERT over PubMedBERT.\\
However, even among the sentence-transformer embedding models, the top performers were the ones that are fully generalists, with jina-embeddings-v2-base-en outperforming by 6\% on the Exact Matching Rate metrics the S-PubMedBERT model, which was initially trained on clinical data and thus can be considered as a partially clinical embedding model. This is consistent with some of the existing literature, which indicates that specialized models, either trained from scratch or through fine-tuning, proved to be less accurate on several tasks than more general and larger models \cite{Nori2023, Singh2023, Liu2023}.\\
This higher sensitivity of specialized models might be explained by the fact that clinical texts contain not only medical terms, but also more general and simpler words that contribute strongly to global text understanding, even in a small sentence context. Building a specialized clinical model (either for generative or embedding purposes) could be improved by including also texts with low medical context (general press articles about medical topics) or even without medical context (any other textual dataset), which existing clinical models did not include in their training or fine-tuning. In addition, generalist models tended to be trained on more data, mainly due to the smaller number of medical textual datasets available compared to more general datasets.\\
This suggests that a good embedding model requires not only a good training procedure (sentence transformer), but also adequate training data, both in terms of quantity and generality, to produce a more robust and accurate model, even in a clinical context.\\

\subsection{Future axis of works}
The proposed study provides solid results and conclusions that can be easily reproduced and extended in two dimensions: width and depth \cite{Clusmann2023}.\\
A broader benchmarking could be done using additional embedding models, especially those that are either pay-to-use models (such as OpenAI's ChatGPTs interface) or much more computationally intensive for inference, which currently limits the reproducibility of our study, hence our choice of embedding models as described in the Methodology section. In addition, a sentence-transformer embedding model that is fully specialized for the clinical domain could be added to this benchmark to accurately compute the gain provided by the sentence-transformer features alone. However, at the time of our study, no fully clinical sentence-transformer embedding model was among the most used HuggingFace models, and S-PubMedBERT was both trained on additional general data and outperformed by fully generalist embedding models. A fully clinical sentence-transformer model could be obtained by training from scratch or using new models that will become available after the publication of this study.\\
Deeper understanding can also be achieved by expanding the context of the input. Larger texts, from paragraphs to entire medical documents such as discharge notes, can be used as a basis for further benchmarking of clinical tasks (summarization, medical code detection, etc.). Such benchmarking and analysis could then be performed in a complete pipeline of a more concrete solution (which could include methods such as RAG systems, for example \cite{Lewis2020}) to quantify whether sentence transformers still perform best in this case, and whether generalist embedding models are still able to be more robust in this scenario of longer medical texts.\\

\textbf{Conclusion}: Generalist embedding models are better than domain specific embedding models in short context clinical semantic search because they have less sensitivity. This may be due to the fact that they have been trained on larger and more diverse datasets, and thus may have a more robust language understanding.\\

\section*{Abbreviations}
The following abbreviations are used in this manuscript:\\

\noindent 
\begin{tabular}{@{}ll}
	CER & Character error rate\\
    ICD-10 & International Classification of Diseases standard v10\\
    ICD-10-CM & International Classification of Diseases standard v10 Clinical Modification\\
    LLM & Large Language Model\\
    RAG & Retrieval Augmented Generation\\
\end{tabular}

\bigskip

\bibliographystyle{unsrt}  
\bibliography{biblio_article}

\end{document}